\documentclass{article}

\usepackage{graphicx}
\usepackage{amsmath}
\usepackage{url}
\usepackage{caption}
\usepackage{subcaption}
\usepackage{float}
\usepackage{booktabs}
\usepackage{microtype}
\usepackage{hyperref}
\usepackage[accepted]{icml2025}
\usepackage[numbers]{natbib}

\begin{document}
\twocolumn[
\icmltitle{Planning vs Reasoning: Ablations to Test Capabilities of LoRA layers}

\begin{icmlauthorlist}
\icmlauthor{Neel Redkar}{UCLA}
\end{icmlauthorlist}

\icmlaffiliation{UCLA}{University of California Los Angeles}

\icmlcorrespondingauthor{Neel Redkar}{neel.redkar@gmail.com}

\vskip 0.3in
]

\printAffiliationsAndNotice{}

\begin{abstract}
Low-Rank Adaptation (LoRA) layers have emerged as a promising approach for efficient model fine-tuning, but their capabilities and limitations remain unexplored. Through systematic ablation studies on GPT-2, we demonstrate that \textbf{reasoning capabilities exist primarily in low-rank spaces} and can be effectively enhanced using LoRA layers, with trained LoRA matrices showing 2-3x lower rank requirements for reasoning tasks compared to planning tasks. We introduce \textbf{HashChain Reasoning}, a novel evaluation dataset for deterministically testing reasoning capabilities, and propose \textit{\textbf{ELoRA}} (Entropy LoRA), a new adapter architecture that improves reasoning ability while speeding up convergence. Our findings are validated through experiments showing that ELoRA performs approximately 5\% better on GSM8k compared to regular LoRAs, along with demonstrating faster convergence rates.
\end{abstract}

\section{Background}

Efficiently handling continual learning for new skills is essential to scale language models \citep{shi_continual_2024}. One current approach that has gained traction in the past year is Low Rank Adaptation (LoRA) layers, which add a couple of low rank linear parameters to efficiently fine tune the model \citep{wistuba_continual_2023}. These parameters can either be applied dynamically or merged into the model via a linear combination. An important question here is the learning capacity of LoRA layers and where the boundaries might be. 

Distinguishing which capabilities LoRA layers are able to fine-tune is key to developing specialized models.

\subsection{Continual Learning}
Continual learning has been a large problem in the LLM space where learning new features is extremely difficult without forgetting the prior distribution. In practice this lends to large decreases in reasoning capability or catastrophic forgetting when fine-tuning models on new tasks. Different solutions have been introduced but each come with their own downsides.

Replay-based methods use a buffer of past actions which can be used in conjunction with fine tuning to make sure the model doesn’t forget key attributes \citep{shi_continual_2024}. With reasoning being an emergent property without any direct training data, this proves to be hard. Regularization is also utilized here but seems to be a half-fix without giving guarantees on knowledge retention and minimizes the effectiveness of finetuning \citep{shi_continual_2024}.

\begin{figure*}[t]
    \centering
    \begin{subfigure}[b]{.48\textwidth}
        \centering
        \includegraphics[width=\textwidth]{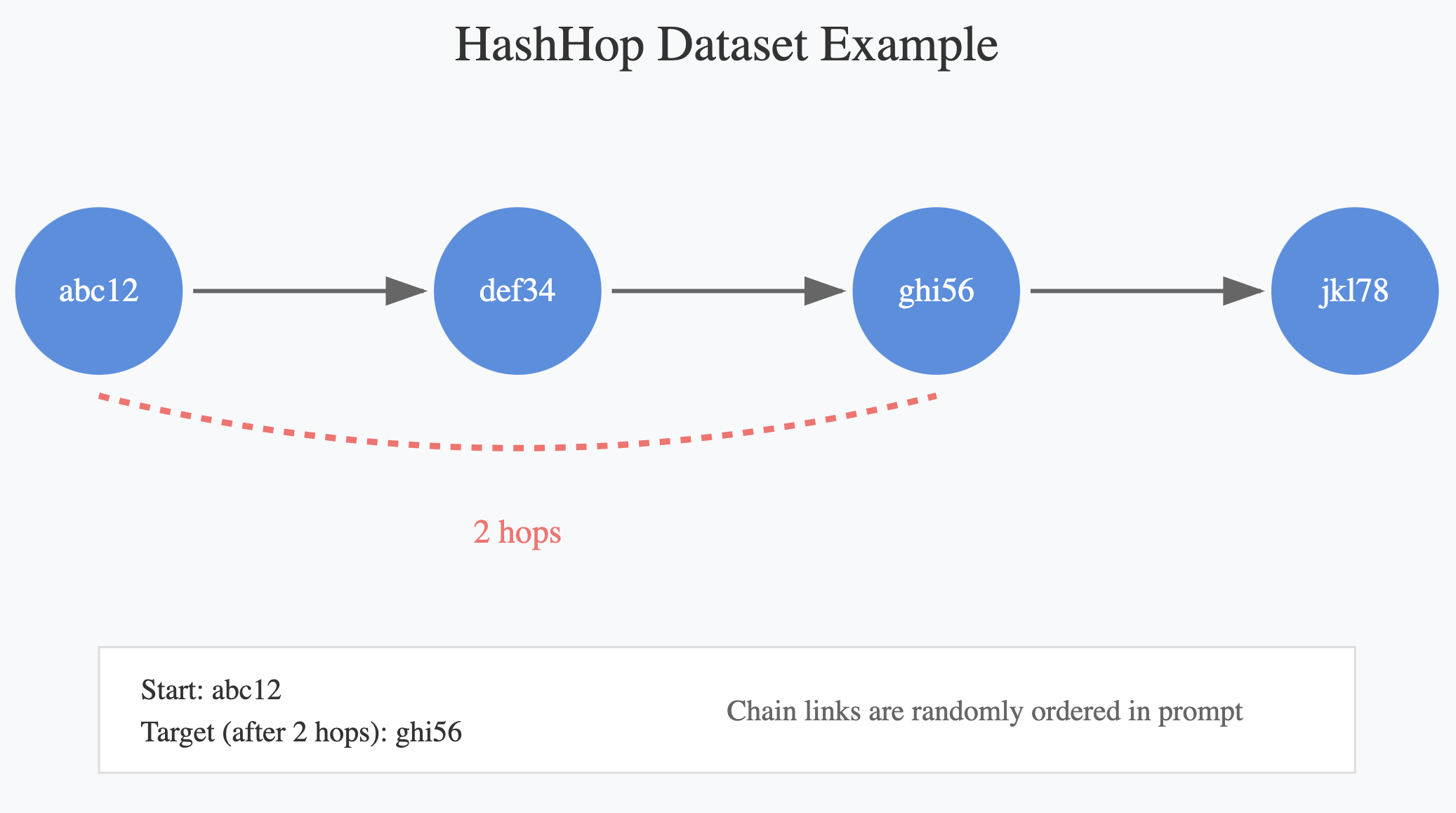}
        \caption{An example of what a hash chain might look like for the regular HashHop task. $abc12$ maps to $ghi56$, because it is 2 arrows ahead.}
        \label{fig:hash-chain}
    \end{subfigure}
    \hfill
    \begin{subfigure}[b]{.48\textwidth}
        \centering
        \includegraphics[width=\textwidth]{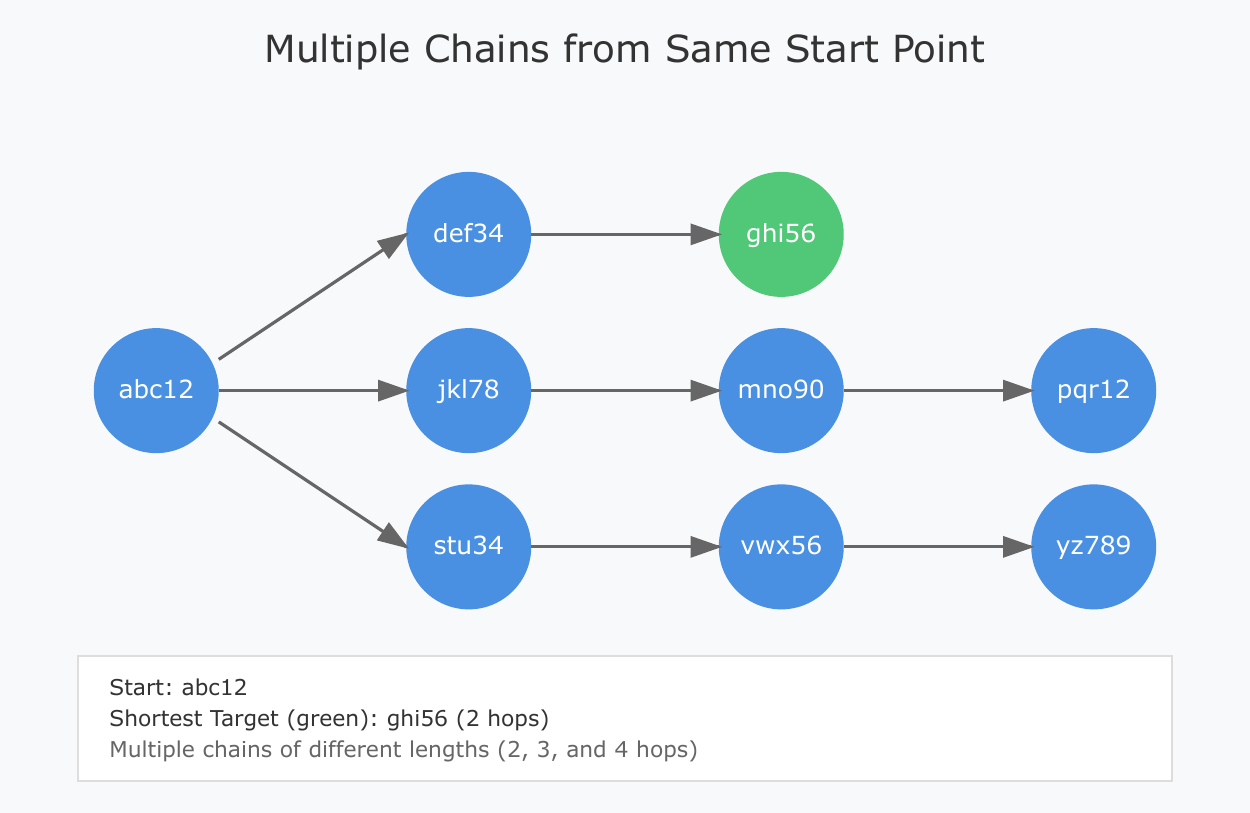}
        \caption{HashChain Reasoning Example: There are 3 chains originating from $abc12$ of length 2, 3, and 3. The ending hash of the shortest chain is $ghi56$ (the correct target).}
        \label{fig:reasoning-hash}
    \end{subfigure}
    \caption{HashHop examples showing (a) a basic hash chain and (b) the structure of the new proposed HashChain Reasoning eval with multiple chains.}
    \label{fig:hash-examples}
\end{figure*}

\subsection{Low-Rank Adaptation (LoRA) Layers}
Low-Rank Adaptation (LoRA) represents a significant advancement in parameter-efficient fine-tuning methodologies \citep{hu_lora:_2021}. By introducing low-rank decomposition matrices into the adaptation process, LoRA enables model specialization while utilizing only a fraction of the parameters typically required for full fine-tuning—replacing an $(n, m)$ matrix with two $(n, rank) \times (rank, m)$ matrices.

This parameter efficiency leads to reduced catastrophic forgetting and minimizes distribution shift during adaptation \citep{biderman_lora_2024}. The approach is particularly compelling from a theoretical perspective, as the constrained parameter space provides implicit regularization against overfitting \citep{wistuba_continual_2023}. This aligns closely with the minimum description length principle-by restricting the dimensionality of possible adaptations, we can identify minimal distribution shifts that achieve task-specific optimization while preserving the model's core capabilities.

Other methods include mixture of expert models, and training new experts for tasks \citep{sukhbaatar_branch-train-mix:_2024}. This seems promising but LoRA layers have benefits in portability, training, and inference capacity \citep{wen_batched_2024}.

It has not been experimented on which specific tasks LoRA layers are useful for and where the boundary exists for when to use LoRA layers versus other parameter efficient methods. This brings up the need to test LoRA layers for specific tasks, especially in the reasoning domain. If the new task requires new reasoning capabilities, are LoRA layers a good fit? Fundamentally this asks the question if reasoning in networks are stored as a low-rank or high-rank construct inside of the weights of the network.

\subsection{Contributions}
\begin{figure*}[t!]
    \centering
    \includegraphics[width=\textwidth]{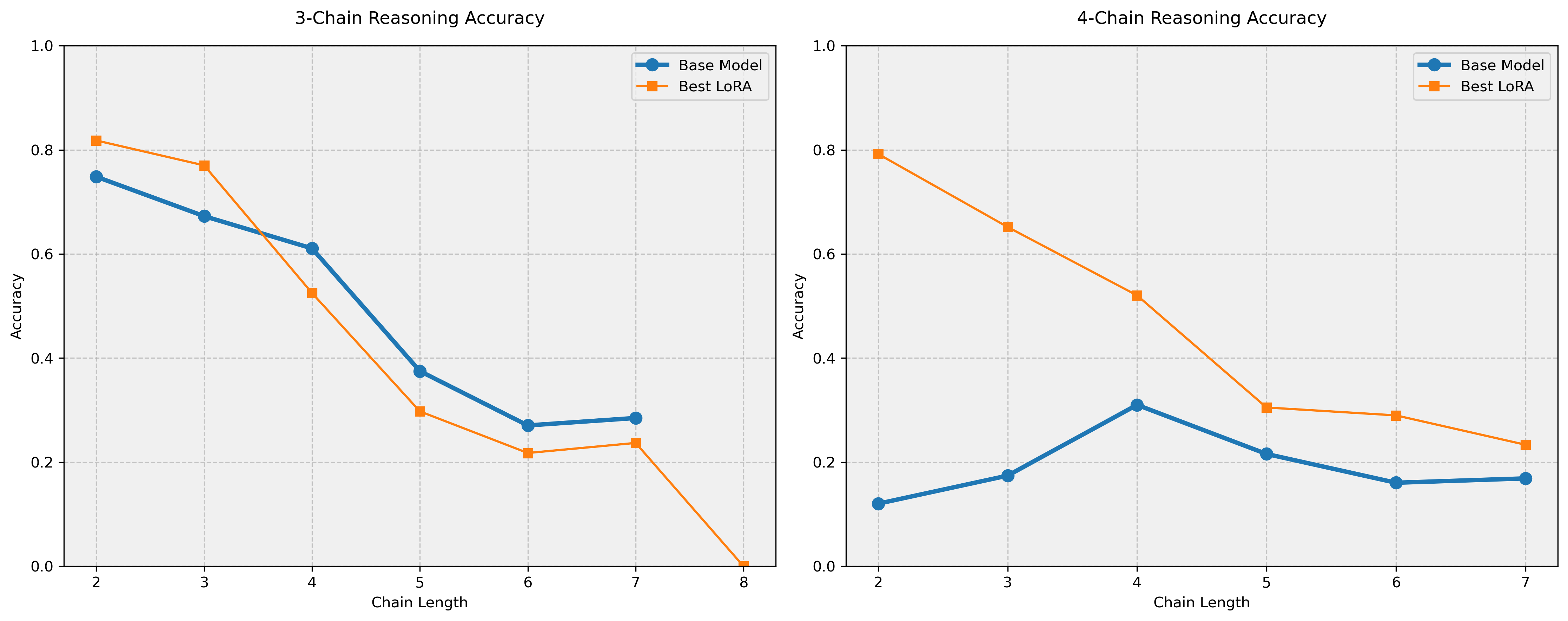}
    \caption{HashChain Reasoning graph with 3 chain (left) and 4 chain (right) accuracies showing performance as individual chain lengths increase. 4 chain reasoning accuracy demonstrated a significant increase with LoRA modules suggesting better reasoning generalization, with minimal fluctuation in 3 chain accuracy}
    \label{fig:chain_reasoning}
\end{figure*}

The contributions that this paper brings are:
\begin{itemize}
  \item A new reasoning eval, \textbf{HashChain Reasoning}, which can be utilized to deterministically check reasoning ability. This is extremely useful for model ablations \& deterministic tests.
  \item \textbf{Reasoning seems to exist low-rank} (Fig. \ref{fig:reasoning_lora_analysis}, \ref{fig:chain_reasoning}), especially compared to planning (Fig. \ref{fig:normal_lora_analysis}). This means LoRA layers could be used for additional reasoning capabilities + might not be as helpful for planning tasks.

  \item \textbf{A new ELoRA (Entropy LoRA) adapter that converges faster and performs better on evals} using a linear approximation for effective rank.  
\end{itemize}

\section{Approach}
For the experimental setup GPT-2 was utilized alongside a custom training script for the LoRA layer addition \citep{radford_language_2019}. First the model is fine-tuned to its capacity on a certain set of reasoning evals (we will get into what these evals are later). Then using LoRA module, we try to increase the model’s capabilities on these evals. Given improvements, this suggests reasoning/"circuits related to reasoning" are low-rank and could be augmented to increase abilities or for cross domain transfer. Further analysis on the LoRA layers themselves could be done to determine the functional rank of the matrices.

Here, reasoning and planning are thought of in terms of tree search algorithms. Planning would be the depth to which the algorithm can search to and reasoning would be the heuristic or communication between branches. Both are needed for more fundamentally complex models.

\subsection{HashHop Dataset}

The first dataset used was HashHop, an eval metric initially created for context window evaluations \citep{hsieh_ruler:_2024} \citep{magic_hashhop:_2024}. A markov chain of hashes is generated and the model is asked to predict \textit{n} hops ahead of the chain. Hashes are utilized because they are completely random. 

The chain is expressed as a randomized list of relations (Fig. \ref{fig:hash-chain}):
\begin{verbatim}
Map:
def34=>ghi56
abc12=>def34
Start: abc12
Hops: 2
Target: [Predicted hash (ghi56) here]
\end{verbatim}

If done in a single token (as in this paper) it shows forward planning ability within the model. This is because rather than doing discrete 1 step jumps at a time, the model has to do \textit{n} jumps in reasoning within a single token.

Due to the model using hashes that are completely non-correlated, this problem cannot be simply solved and the model needs to do one hop at a time. A high accuracy would guarantee that the model has gone through all the hashes in the chain.

The dataset is deterministic and has reasonable analogs in common tasks such as linking together events. Since the metric is also dynamically generated, it is impossible to “overfit” to the eval like other tasks.

This model for reasoning has a couple shortcomings though, where the benchmark is relatively artificial due to its usage of hashes. The utilization of hops as a corollary for reasoning is also not common practice and might not necessarily reflect reasoning more-so as planning.

Another approach to thinking about this task is looking at the chain of thought solution. The model could much more easily solve this problem in small steps (manually drawing out $a \xrightarrow[]{} b \xrightarrow[]{} c \xrightarrow[]{} d$ etc.). By compacting the task into a single token, we now have a metric for ``planning ahead" or functionally skipping reasoning steps, going straight for the solution (predicting \textit{n} hashes ahead).
\begin{figure*}[t!]
    \centering
    \includegraphics[width=\textwidth]{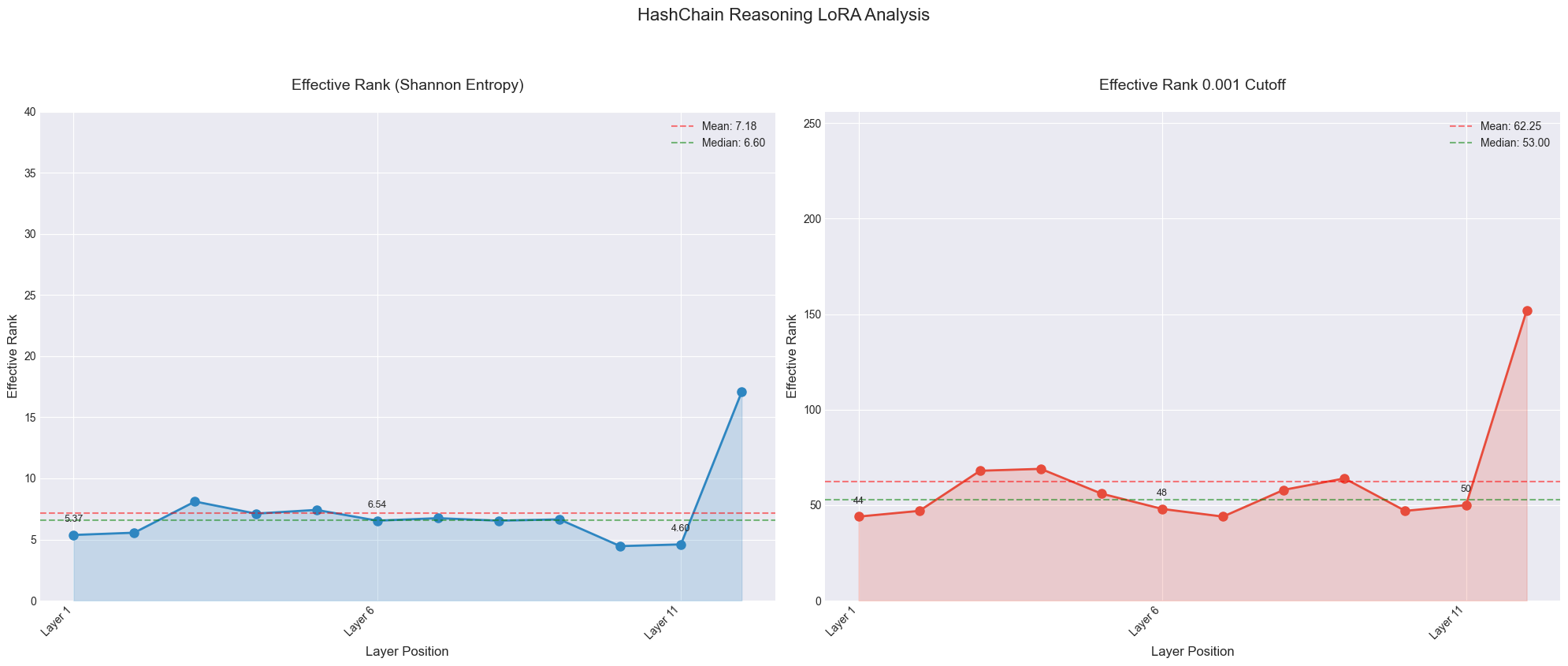}
    \caption{Using the same metrics, Shannon entropy \citep{roy_effective_2007} (left) and cutoffs (right), there is about a 2x-3x decrease in effective rank for the HashChain reasoning task compared to the baseline HashHop task.}
    \label{fig:reasoning_lora_analysis}
\end{figure*}
\subsection{HashChain Reasoning Dataset}

This dataset utilizes similar concepts from the HashHop eval but instead has multiple chains of varying lengths \ref{fig:reasoning-hash}. The model is then asked to predict the hash of the shortest chain. The eval is still artificial in nature but arguably more representative of common reasoning tasks in practical tasks.

The HashChain template is:

\begin{verbatim}
Map:
abc=>mno
abc=>ghi
abc=>jkl
ghi=>xoh
jkl=>djw
Start: abc
Task: shortest path
Target: [Predicted hash (mno) here]
\end{verbatim}

If one thinks of reasoning as a breadth-first tree search, HashChain effectively horizontally scales. This is opposed to HashHop vertically scaling the search in a planning dimension. This analogy makes sense for the dataset as the optimal way solve HashChain is through a breadth-first search.

Specifically for reasoning, take common hard tasks such as “Which number is greater, 2.11 or 2.9?” or other logical tasks. These tasks fundamentally need comparison of multiple branches/sections. For the numerical example it would be “2.11” vs “2.9” but for more abstract reasoning it might be asked to perform boolean logic (AND, OR, NOT, etc.) with multiple branch comparisons.

HashChain Reasoning would cover this by guaranteeing that a “comparison operation” or heuristic between multiple branches would have to be done to find the correct hash. This is explicitly different from HashHop as the model only needs to hold one thread.
\subsection{Training}

The goal is to fully finetune a small model on the eval dataset and see if through LoRA layers our metrics increase in a substantial way. The hyper-parameters were fixed and iterations were done till the loss plateaued at an equilibrium. All training scripts are open source \href{https://github.com}{https://github.com/anon/opensource}.


\section{Experimental Results}

\begin{table}[t]
\centering
\caption{Averaged Model Accuracy Comparison}
\label{tab:model-comparison}
\begin{tabular}{lccc}
\toprule
\textbf{Model} & \textbf{Base} & \textbf{LoRA} & \textbf{ELoRA} \\
\midrule
HashHop & 0.283 & 0.302 & 0.303 \\
Reasoning-3chain & 0.391 & 0.452 & 0.473 \\
Reasoning-4chain & 0.192 & 0.369 & 0.451 \\
\bottomrule
\end{tabular}
\end{table}

\begin{figure*}[t!]
    \centering
    \textbf{HashHop Regular LoRA Analysis}
    \includegraphics[width=\textwidth]{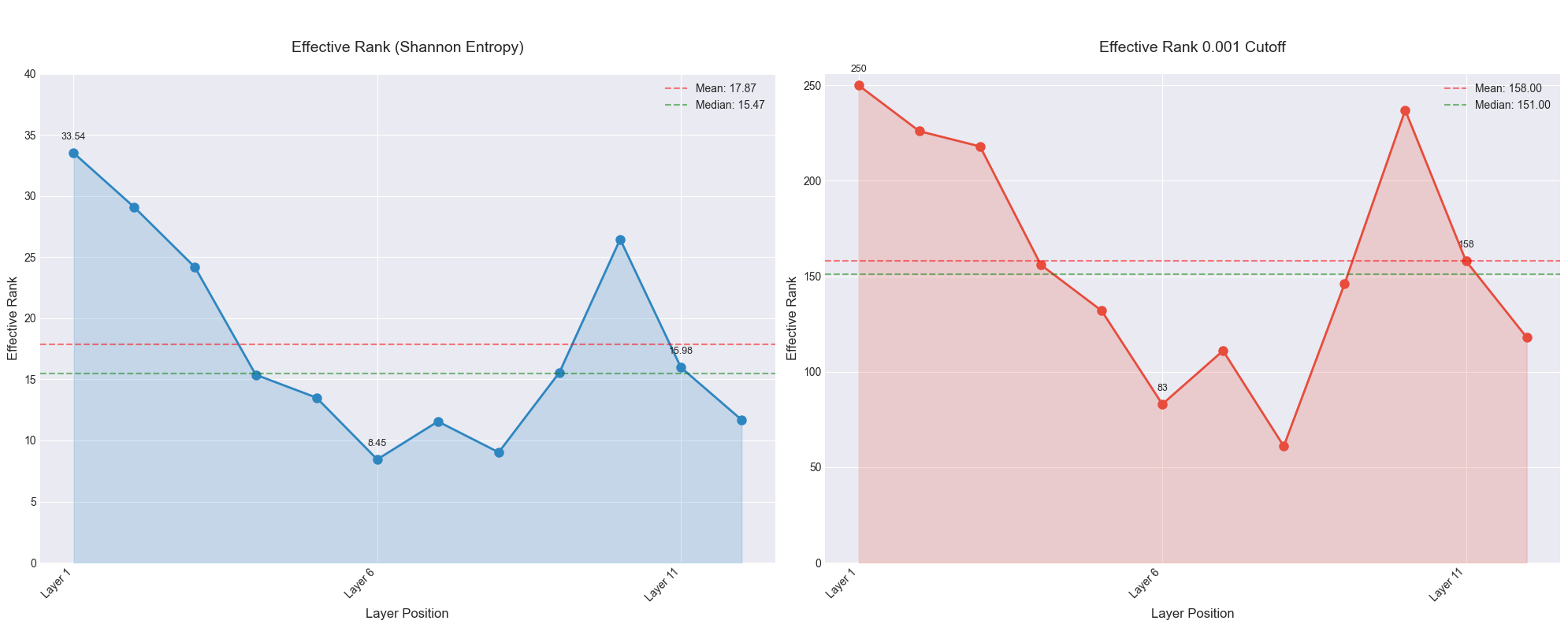}
    \caption{Calculating the effective rank of the LoRA matrix trained on the regular HashHop task. Shannon Entropy \citep{roy_effective_2007}(left) gave a mean of 17.87 and a cutoff (right) gave a mean of 158}
    \label{fig:normal_lora_analysis}
\end{figure*}

\subsection{HashHop Evaluations}

\begin{figure*}[t]
    \centering
    \begin{subfigure}{.48\textwidth}
        \includegraphics[width=\linewidth]{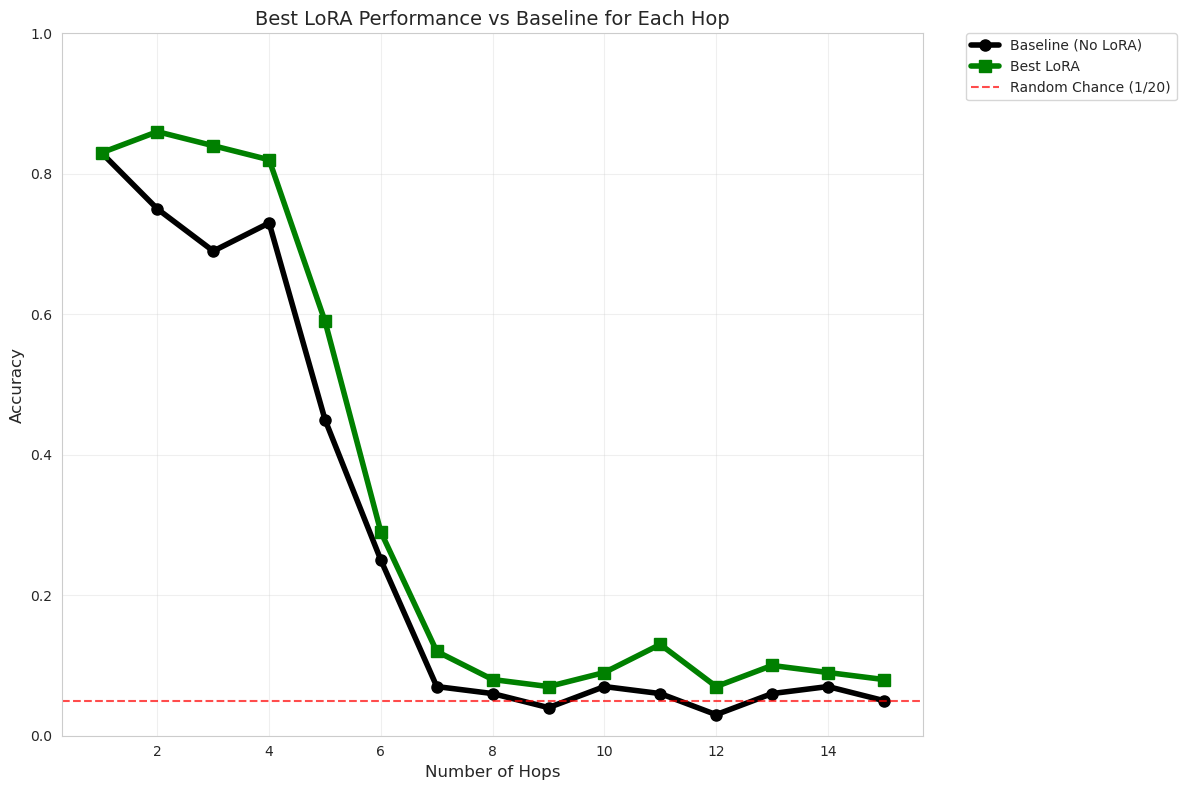}
        \caption{Initial accuracies of finetuning GPT2 on HashHop (black) and subsequent addition of LoRA layers (green) }
        \label{fig:perf_comparison}
    \end{subfigure}
    \hfill
    \begin{subfigure}{.48\textwidth}
        \includegraphics[width=\linewidth]{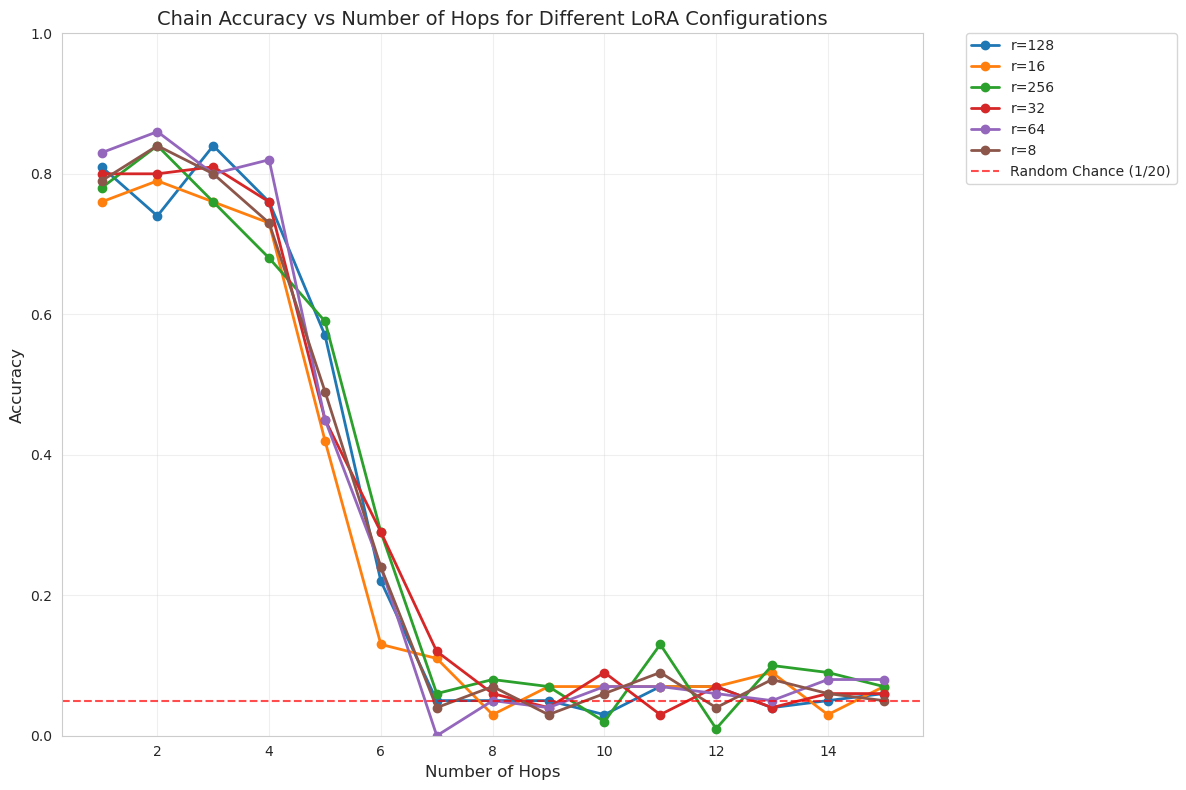}
        \caption{General LoRA ablations for HashHop}
        \label{fig:lora_ablations}
    \end{subfigure}
    \caption{Negative result for regular HashHop LoRA tests. LoRA layers only increase previously known planning capabilities, without any large boosts in accuracy.}
    \label{fig:hashhop_analysis}
\end{figure*}

The chain length was varied between 1-20, with 15-20 being cut off in the graph for being equally close to random chance. For the regular HashHop evaluations the prediction power increased marginally with LoRA layers within the first couple hops \ref{fig:perf_comparison}. Jumps in accuracy started larger, ~10\% for 2 hop, but quickly deteriorated to random chance.

The curves all seem similar to a sigmoid function which start high around ~80\% accuracy and quickly drop to random chance. The quick drop highlights the difficulty of the task and its usefulness to benchmark emergent planning abilities. LoRA layers were also selectively fine-tuned on 4-20 hops and the same results were found.

\begin{figure}[t!]
\centering
\textbf{ELoRA Architecture}
\includegraphics[width=0.9\columnwidth]{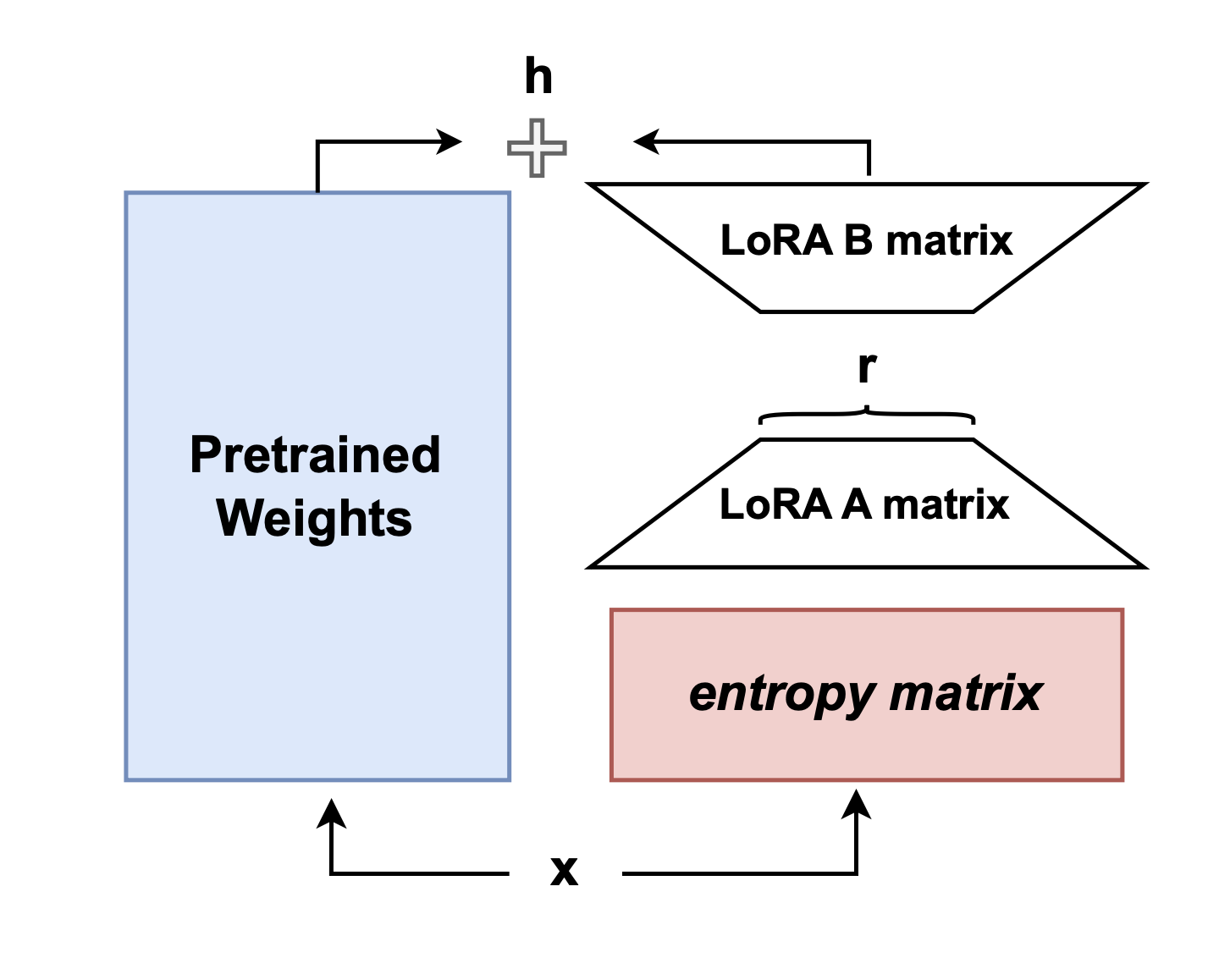}
\caption{ELoRA architecture diagram. Similar to LoRA except for a small linear matrix with maximized entropy.}
\label{fig:elora-arch}
\end{figure}

This suggests that although LoRA layers are helpful for fine-tuning specific aspects like style, it is unable to teach fundamentally new planning capabilities to the network. The jumps in accuracy (2, 3, 4, \& 5 hop) were on tasks that the model had already expressed prior functionality in. In Figure \ref{fig:lora_ablations} and \ref{fig:normal_lora_analysis}, ablations were done on the rank of the resultant LoRA matrix. The effective rank of each LoRA matrix seems to hover around 150, suggesting that the task is hard to learn and high rank \citep{roy_effective_2007}.

The result is also supported in the literature. Due to LLM's being next token predictors, long term planning is especially hard \citep{bachmann_pitfalls_2024}. If thought of in the same way as the breadth-first HashChain, this proves that vertically scaling reasoning is fundamentally high rank/difficult.

\subsection{HashChain Reasoning Evaluations}
The model was finetuned to capacity on 3-chain and 4-chain accuracy. The model converged on learning to perform well on 3 chains and seemed to perform less accurately on 4 chains as seen in Figure \ref{fig:chain_reasoning}. Chain length in the reasoning tasks seemed to have much less of an effect on accuracy than in the regular HashHop evaluations, where the accuracy degraded linearly instead of exponentially as before.

Similar to planning, LoRA modules were trained on weak areas for the model. In 4-chain accuracy we see a huge boost due to the LoRA module, while having minimal degraded accuracy with the 3-chain accuracy. This shows that the model was able to increase its capabilities from being unable to perform 4-chain reasoning to gaining an 80\% accuracy with lengths of 2 and genrally significantly higher accuracies than before.

The effective rank of the LoRA layers was also found to be significantly lower than the regular eval, as shown in Figure \ref{fig:reasoning_lora_analysis}. LoRA rank hovered around 50 for most layers, except the last. An effective rank of 50 is ~3x less than the parameters needed for the regular HashHop model, suggesting a fundamental difference in the learned data distributions, with the HashChain Reasoning dataset being simpler to represent in latent space.

This shows the horizontal scaling of reasoning or handling multiple threads of reasoning is low rank, and can be expanded with LoRA layers. The implications are large as any reasoning task that can be formalized into concurrent operations can then be improved via LoRA layers (retrieval tasks, knowledge addition, or arithmetic with multiple numbers)

Spiking near the end for both could also be attributed to latter layers often needing more LoRA parameters \citep{gao_higher_2024}.

\section{ELoRA Validation Tests}

\begin{figure}[t!]
\centering
\textbf{ELoRA Convergence}
\includegraphics[width=0.9\columnwidth]{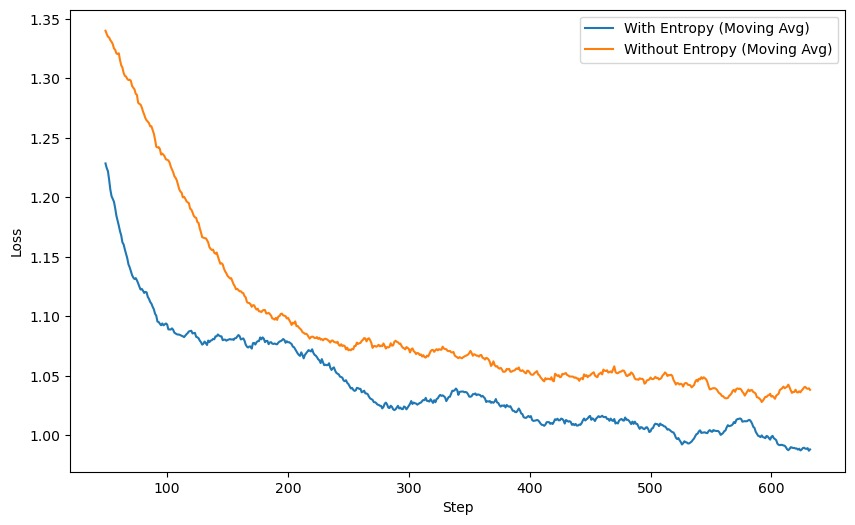}
\caption{ELoRA converges faster than LoRA on average when given the same data}
\label{fig:elora-converge}
\end{figure}

There have been many variations to the LoRA architecture, such as taking weight decompositions like DoRA (Weight-Decomposed Low-Rank Adaptation) and QLoRA (Quantized Low Rank Adaptation) both for reasoning and efficiency gains respectively. It has been hard to find a good variation for LoRA's that converge fast as well as have good reasoning priors.

Here we introduce ELoRA (an entropy LoRA) to validate our analysis in the prior section \ref{fig:elora-arch}. By having a linear layer prior to the LoRA that maximizes entropy and disentangles the representations, we can utilize the ability of LoRAs to effectively "pick out" specific representations into a low rank space. This method has been utilized before to prevent representation collapse. \citep{arefin_seq-vcr:_2024} \citep{skean_frossl:_2024}

The ELoRA is structured similarly to a regular LoRA but with an entropy matrix beforehand with a modified loss term. There is no activation function, which allows the adapter to be merged into the weights after training. Learning rates were set to $1e-3$ for the matrix and $1e-5$ for the regular low-rank matrices and Phi1.5b was utilized for training. \citep{li_textbooks_2023}

The use of an external matrix could also allow for faster transfer learning convergence through a learned "high entropy prior".

For the loss, we utilize a linear entropy equation coined in FroSSL that is equivalent to Rényi $\alpha$-entropy where $\alpha=2$ \citep{skean_frossl:_2024}. The benefit is that one does not have to calculate the singular values, which is expensive for large matrices. The loss term of the entropy is given below:

$$\mathcal{L}_{\text{entropy}} = \log(|Z^TZ|_F^2)$$

\begin{table}[t]
\centering
\caption{GSM8K Validation Accuracy by Adapter}
\label{tab:gsm8k-results}
\begin{tabular}{lc}
\toprule
\textbf{Adapter} & \textbf{GSM8K Validation} \\
\midrule
Phi 1.5 base & 33\% \\
LoRA & 43\% \\
ELoRA & 48\% \\
\bottomrule
\end{tabular}
\end{table}

The results of the experiment were promising and seem to validate the prior sections. ELoRA converged faster than regular LoRAs \ref{fig:elora-converge} and consistently scored higher ~5-6\% higher in GSM8k math evaluations \citep{cobbe_training_2021} \ref{tab:gsm8k-results}. ELoRA also consistently outperforms LoRAs in HashHop but much more significantly in HashChain where more reasoning is needed \ref{tab:model-comparison}.

\section{Conclusion}

LoRA layers seem to be the solution for increasing reasoning capabilities of models and it it might even be possible to train “reasoning modules” on key datasets to scale broad range reasoning ability. The same does not seem likely for planning due to its higher rank and inability to be boosted in new capacities. Splitting evals into these two categories and having granular objectives is important to decrease ambiguity (ie. differentiate high reasoning with low planning or vice versa).

The introduction of ELoRA provides additional evidence for these observations. On the GSM8k benchmark, ELoRA showed a modest but consistent improvement (~5\%) over standard LoRA implementations, with notably faster convergence rates. The entropy component appears to aid in representation learning, by disentangling the representations.

The 4-chain abilities in the HashChain Reasoning dataset saw large increases in accuracy in regular LoRA's and ELoRA's, suggesting not just improving reasoning but also horizontally adding new capabilities. Distinguishing between planning \& reasoning is necessary because as we add LoRA layers, its important to check whether or not LoRA layers are the right fit for the job—especially with continual learning.

Planning ability did increase but only in areas that the model already excelled in. This suggests LoRA layers could be used to ``squeeze out" more latent planning ability but planning still seems fundamentally hard for LoRA layers \citep{bachmann_pitfalls_2024-1}.

Our current intuition for why reasoning could be low-rank is that a ``simplicity prior" is important to generalization. Following the minimum description length principle, having simple circuits for higher order operators might allow for generalization across other higher order concepts in the model.

\subsection{Future Research Directions}
The use of effective rank + targeted datasets to check the difficulty of tasks for models to learn seems promising, as well as use of LoRA layers to check effective rank of circuits in models. LoRA layers could possibly be used as a replacement for Mixture of Experts or other techniques for increasing parameter counts in a sparse way for reasoning tasks \citep{sukhbaatar_branch-train-mix:_2024}. 

The promising initial results of ELoRA help the reasoning claims in the paper. More comprehensive benchmarking across a broader range of reasoning tasks (such as BIG-Bench, MMLU, and BBH) would help establish the generality of ELoRA's benefits beyond GSM8k and HashChain Reasoning. Of particular interest is the potential use of the learned entropy matrix as a transferable prior—preliminary results suggest faster convergence.

Testing these benchmarks with new reasoning models such as Deepseek R1 seems interesting, as intermediary layers seem to contain higher entropy than non CoT distilled counterparts \citep{deepseek-ai_deepseek-r1:_2025}.

Utilization of LoRA features this way lends itself well to continual learning \& might provide a way to combat forgetting. By using task-specific LoRA layers, a model can continually adapt to new distributions—possibly even mixing them together. The main unsolved problem would be training a sufficiently complex router that can be frequently updated.

LoRA layers could also be used to increase reasoning in domain specific areas (ie. comparison for retrieval, or math for technical domains). The layers could then be merged back into the model to have a base model better at reasoning than before. If one imagines world models stored as graphs, then it should be possible to horizontally increase the capability of the model to ``process" this graph.

Separating planning and reasoning also brings up the interesting follow-up where if planning is hard for language models, what changes to the structure of transformers could allow for greater planning ability? Would it also be unfair to test new planning techniques against reasoning baselines (It seems easy to imagine a world where one can plan hundreds of steps ahead, but be limited by its ability to reason between optimal outcomes)?

Code + benchmarks are public and open source @ \href{https://github.com/anon/opensource}{https://github.com/anon/opensource}
\bibliographystyle{plainnat}  
\bibliography{citations}

\end{document}